\documentclass[10pt,letterpaper]{article}

\usepackage{cogsci}

\cogscifinalcopy 

\usepackage{multirow}
\usepackage{color}
\usepackage{verbatim}
\usepackage{makecell}
\usepackage{pslatex}
\usepackage{apacite}
\usepackage{float} 
\usepackage{graphicx}
\usepackage{array}
\usepackage{caption}
\usepackage{mdframed}
\usepackage{paralist} 
\usepackage{threeparttable}

\title{``There Is No Such Thing as a Dumb Question," But There Are Good Ones}

\author{{\large \bf Minjung Shin (mjshin77@snu.ac.kr)} \\
  Interdisciplinary Program in Cognitive Science, Seoul National University, Seoul, 08826, Republic of Korea\\
  \AND {\large \bf Donghyun Kim (dhyun0704@kaist.ac.kr)} \\
  Department of Brain and Cognitive Sciences, KAIST, Daejeon, 34141, Republic of Korea\\
  \AND {\large \bf Jeh-Kwang Ryu (ryujk@dgu.ac.kr)} \\
  Department of Physical Education, Dongguk University, Seoul, 08826, Republic of Korea}

\begin{document}

\maketitle

\begin{abstract}
Questioning has become increasingly crucial for both humans and artificial intelligence, yet there remains limited research comprehensively assessing question quality. In response, this study defines good questions and presents a systematic evaluation framework. We propose two key evaluation dimensions: appropriateness (sociolinguistic competence in context) and effectiveness (strategic competence in goal achievement). Based on these foundational dimensions, a rubric-based scoring system was developed. By incorporating dynamic contextual variables, our evaluation framework achieves structure and flexibility through semi-adaptive criteria. The methodology was validated using the CAUS and SQUARE datasets, demonstrating the ability of the framework to access both well-formed and problematic questions while adapting to varied contexts. As we establish a flexible and comprehensive framework for question evaluation, this study takes a significant step toward integrating questioning behavior with structured analytical methods grounded in the intrinsic nature of questioning.

\textbf{Keywords:} 
Question quality evaluation; Artificial Intelligence(AI); Large Language Model(LLM); Rubric
\end{abstract}

\section {Introduction}

Questioning is prevalent in all forms of discourse; however, it remains elusive in systematic approaches \cite{graesser1985psychology}. Instead, there is a pervasive tendency to encourage questioning itself rather than evaluating its quality \cite{flammer1981towards}, as exemplified by Carl Sagan's famous quote, ``There is no such thing as a dumb question." \cite{sagan2011demon}

Although questioning has been a cornerstone of human discourse throughout history, it has only recently emerged as a crucial element in artificial intelligence(AI)  \cite{shin2023uncertainty}. Despite their remarkable generative capabilities, current AI systems struggle with real-world uncertainty and change \cite{marcus2020gpt}. Given the growing demand for human-AI interaction, questions offer a simple yet powerful tool for addressing the inherent ambiguity of human language \cite{bender2020climbing}. This is particularly crucial, as AI systems' probabilistic generation can pose risks in critical domains \cite{amodei2016concrete}, suggesting a transition toward more deliberative validation steps instead of jumping to conclusions \cite{toles2023good}.
These insights suggest a paradigm shift in AI development from fluent output generation toward question-centric approaches.

However, there is a notable scarcity of studies on `how to ask' compared to on `how to answer' in both human-centered and AI-focused research. This research deficit translates into significant gaps in both the systematic analysis and evaluation of questioning \cite{Graesser2009whatGQ}. 

Such scarcity can be attributed to two factors: 1) encouraging questioning instead of judging it and, 2) the unique role of questions in language use -- questions are not a proposition but a \textit{speech act}\footnote{a linguistic actions fulfilling social functions, like promising, commanding, or questioning, rather than just conveying information} aimed at obtaining information \cite{Huddleston2002ClauseType}. Questions do not have absolute truth values \cite{searle1969speech, graesser1985introduction}, so there is inherently no `right' or `wrong' way to ask a question. 

Extending the context to daily interactions makes `apt questioning' more elusive. Questions shape conversation flows \cite{graesser1985introduction}, build social rapport \cite{kim2022prosocialdialog}, and induce reflective thinking \cite{petty1981effects}. As context deepens, the question appropriateness depends on various paralinguistic elements, making assessing question quality in absolute terms challenging. As a result, determining ``Is it a good question?" is more difficult than ``Is it a proper answer?"

\begin{table*}[ht!]
    \centering
    \caption{Key Criteria for Question Evaluation}
    \label{tab:question_eval_basic}
    \renewcommand{\arraystretch}{1.5}%
    \begin{tabular}{{p{0.12\textwidth}  p{0.4\textwidth}  p{0.4\textwidth}}}
    \hline
         & \textbf{Appropriateness} & \textbf{Effectiveness}\\
    \hline
        Definition & \textbf{Sociolinguistic competence} to use language flexibly according to given situations and contexts without violating behavioral standards or norms & \textbf{Strategic language competence} that sucessfully fulfills individual intentions and accomplishing desired outcomes or goals.\\
        Focus & Context-based evaluation & Goal achievement-centered evaluation \\
        Key question & Is the question aligned with the flow and context of the discourse? & Does the question achieve its intended purpose? \\
        Characteristics & Can score high even if not aligned with purpose & Must align with purpose to score \\
    \hline
    \end{tabular}
    \label{tab:GQcriteria}
\end{table*}

Whereas humans can intuitively navigate this complexity without much pressure or demands, machines require more explicit and structured guidance to process the same information. Specifically, machines face fundamental limitations; they rely on probabilistic generation rather than true understanding \cite{marcus2020gpt}, lack social learning abilities \cite{lake2017building}, and have limited pragmatic schema \cite{Bender2021parrots}. Therefore, clear criteria are necessary to systematically analyze and evaluate machine-generated questions.

This study makes three main contributions to question evaluation. First, we established evaluation criteria that reflect the intrinsic nature of questioning as a speech act, considering both pragmatic functions and contextual dependencies. Second, we develop a systematic scoring framework using a rubric-based approach. Finally, we implemented an automated evaluation system utilizing Large Language Models(LLMs), demonstrating how structured human-defined criteria can be effectively applied. These contributions advance our understanding of question quality assessment and provide practical tools for improving human-AI interactions through better Question Generation(QG) and evaluation.

\subsection{Question Evaluation in NLP}
 
Researchers in Natural Language Processing(NLP) proposed new QG evaluation methods to highlight the shortcomings of traditional techniques \cite{mulla2023automatic}. The conventional method relies on sentence similarity metrics to compare model-generated questions with human-written reference sentences \cite{papineni2002bleu, lin2004rouge, banerjee2005meteor}. While this approach is effective for other NLP tasks, such as translation or summarization, it does not align with question's context-dependency and one-to-many nature.

Some researchers address these challenges by suggesting multiple reference sentences \cite{oh2023evaluation} or unsupervised methods \cite{ji2022qascore}, due to the practical impossibility of obtaining diverse reference sentences given the open-ended nature of questions. \cite{qi2020stay}. Human evaluation offers an alternative through criteria-based scoring of fluency, relevance, accuracy, and difficulty \cite{mulla2023automatic}, but remains cost-inefficient and vulnerable to subjective bias.

Amidst these limitations, domain-constrained evaluation criteria have arisen in various fields. Machine reading comprehension systems focus on question diversity \cite{sultan2020diverseQG, yoon2023storybook}, visual QG systems emphasize uniqueness \cite{Jain2017CVPR}, and the Questioning Turing Test framework considers human-likeness, correctness, and strategicness \cite{damassino2020QTT}. However, these approaches remain fragmented and limited in scope.

The fundamental diversity of questions intensifies challenges when extended to real-world contexts. Questions vary in form and type, with evaluation needs differing by conversational purpose \cite{flammer1981towards, graesser1992mechanisms}. For instance, clear and concise questions suit collaborative tasks \cite{Rothe2017NIPS, rothe2018people}, while deep and exploratory questions are more effective in educational settings for facilitating student thinking \cite{rickards1974type, dillon2006effect}.

Taken together, existing attempts to evaluate machinery QG show major limitations. Current methods rely on irrelevant automatic metrics, subjective assessments, and rigidly task-specific criteria. These issues hinder high-performance AI in real-world applications, especially in open-ended conversations. Thus, comprehensive evaluation criteria are needed to effectively capture questioning behavior and ensure broad applicability.

\section{Development and Validation of Question Evaluation Metric} \label{sec:GQ_criteria} 

While current LLMs demonstrate human-like language, their fundamentally different mechanisms necessitate a more nuanced comparison. Following  \citeNP{mahowald2024dissociating}, we distinguish between \textbf{formal linguistic competence} (\textit{knowledge of language rules and patterns}) and \textbf{functional linguistic competence} (\textit{the ability to use language in the real world}) to better frame this comparison.

Formal competence involves generating grammatically correct sentences based on language rules and is tied to traditional linguistics. Conversely, functional competence pertains to understanding and using language in real contexts and is closely linked to pragmatics. Both competencies are vital for effective communication, requiring grammatically meaningful utterances and strategic use depending on the situation.

To establish evaluation scope and criteria, we deliberately focused on functional competence. Conventional QG models needed to evaluate whether `the sentence is properly formed,' indicating grammatical correctness \cite{mulla2023automatic}. But current LLMs show considerable achievement in formal linguistic abilities while remaining limited in functional linguistic abilities  \cite{mahowald2024dissociating}.

Overall, this evaluation research delves into pragmatics and social linguistics to assess LLMs' functional competence \cite{graesser1997discourse}. Considering interactive, information-seeking nature of the questions, two key evaluation dimensions emerged from this investigation. These are \textbf{appropriateness}, which evaluates whether a question fits the context, and \textbf{effectiveness}, which focuses on the goal achievement as detailed in table~\ref{tab:question_eval_basic} \cite{spitzberg1994competence, canale2014communicative}. In other words, \textbf{a good question is both appropriate and effective.}

\subsection{Metric Development}
\vspace{-0.15cm}
Questions are linked to their discourse context, shaping structure and coherence. They fulfill various roles beyond information requests, like controlling flow, facilitating topic shifts, and directing listeners' attention. This multi-faceted nature emphasizes the need for nuanced evaluation. Drawing from these characteristics, we identified the following evaluative sub-components of \textbf{appropriateness} and \textbf{effectiveness}.

\subsubsection{Sub-Components of Appropriateness} 
{\setlength{\itemsep}{0pt}
\begin{compactitem}   
    \item \textbf{Cohesion}: Evaluate whether the generated question flows smoothly within the context with proper cohesive markers. Questions that disrupt conversation because of ambiguous elements are regarded as having low cohesion.       
    \item \textbf{Answerability}: Indicates whether a meaningful answer is possible based on available knowledge. Incomprehensible or overly difficult question have low answerability.     
    \item \textbf{Respectfulness}: Evaluates if questions show respect for the other party, considering their feelings and position to build trust and maintain positivity. Rude or aggressive questions lack respect. 
\end{compactitem}}

\subsubsection{Sub-Components of Effectiveness} 
{\setlength{\itemsep}{0pt}
\begin{compactitem}    
    \item \textbf{Clarity}: Relates to how clearly the questioner's goal is understood, affecting response accuracy, efficiency, and understanding. Questions with ambiguous phrasing or vague intention lacks clarity.
    \item \textbf{Coherence}: Refers to how well components connect to create meaningful content, improving the chances of advancing conversation. Logically inconsistent questions or those straying from the purpose receive a low score.     
    \item \textbf{Informativeness}: Indicates how well a question elicits useful information, determining if it can gather relevant data for its goal. Questions about resolved issues or irrelevant topics are deemed low in informativeness. 
\end{compactitem}}

\subsection{Rubric-Based Scoring System}
\vspace{-0.1cm}
\begin{table*}[ht!]
    \centering
    \caption{Evaluation Rubrics for Appropriateness and Effectiveness}
    \label{tab:rubric}
    \vspace{-0.2cm}
    \renewcommand{\arraystretch}{1.5}%
    \begin{tabular}{p{0.5\textwidth}  p{0.45\textwidth}}
    \hline
    \textbf{Rubric for Appropriateness} & \textbf{Rubric for Effectiveness} \\
    \hline
    {\small
    \textbf{[Cohesion]} \newline
    1: Contextually misused cohesive markers. May disrupt conversation \newline
    2: Ambiguous cohesive markers with partial context. Hinders conversation \newline
    3: Adequate cohesive markers with context. Maintains conversation \newline
    4: Well-contextualized cohesive markers. Natural conversation flow \newline
    5: Perfectly contextualized cohesive markers. Very natural conversation flow \newline
    \textbf{[Answerability]} \newline
    1: Unclear or speculative. Impossible for \$\{answerer\} to answer \newline
    2: Somewhat ambiguous or difficult. Challenging for \$\{answerer\} to answer \newline
    3: Generally valid. Some difficulty for \$\{answerer\} to answer \newline
    4: Clear and appropriate. Easy for \$\{answerer\} to answer \newline
    5: Very clear and appropriate. \$\{answerer\} can answer immediately\newline
    \textbf{[Respectfulness]} \newline
    1: Rude and aggressive without consideration for others. Damages atmosphere \newline
    2: Somewhat rude, inconsiderate of others. Negative impact \newline
    3: Generally respectful but needs refinement. Neutral atmosphere \newline
    4: Respectful and considerate of others. Positive atmosphere \newline
    5: Highly respectful and considerate of others. Excellent atmosphere} &
    {\small
    \textbf{[Clarity]} \newline
    1: Unclear structure making \$\{goal\} intent impossible to grasp \newline
    2: Vague structure making \$\{goal\} intent difficult to grasp \newline
    3: Generally clear with some ambiguity in \$\{goal\} intent \newline
    4: Clear structure with easily understood \$\{goal\} intent \newline
    5: Very clear structure with perfectly conveyed \$\{goal\} intent \newline
    \textbf{[Coherence]} \newline
    1: Irrelevant to topic with unclear purpose. Disrupts logical flow \newline
    2: Partially relevant with unclear purpose. Hinders logical flow \newline
    3: Generally relevant with clear purpose. Maintains logical flow \newline
    4: Well-connected to topic and purpose. Natural logical flow \newline
    5: Perfectly relevant and purposeful. Excellent logical flow \newline
    \textbf{[Informativeness]} \newline
    1: Seeks irrelevant or speculative information, hindering \$\{goal\} \newline
    2: Seeks low-relevance information, making \$\{goal\} difficult \newline
    3: Shows potential for \$\{goal\} \newline
    4: Shows high potential for \$\{goal\} \newline
    5: Guarantees \$\{goal\} }\\
    \hline
    \end{tabular}
\end{table*}

We adopted the \textbf{rubric} as a detailed scoring guideline for automatic evaluation. A rubric is an explicit set of criteria developed in education for formative assessment\footnote{An evaluation conducted during the learning process, aimed at diagnosing learners' current state, identifying areas for improvement, and promoting learning.} for unstructured performance. A rubric, comprises \textbf{evaluation criteria}, \textbf{scores}, and \textbf{descriptions} for each score. It offers clear, objective assessments compared to conventional methods, ensuring evaluator consistency \cite{Brookhart2018rubric}.

Applying rubrics to LLMs, as a variant of the LLM-as-a-judge paradigm \cite{li2024llmjudge}, leverages their pattern recognition and context-learning capabilities by incorporating detailed evaluation criteria into prompts. This approach decomposes the evaluation process into multifaceted criteria across multiple dimensions, enabling objective assessment based on human-defined standards while maintaining systematic consistency.

\subsubsection{Setting Scores and Descriptions}
Following established practices in rubric development \cite{popham1997rubric3-5}, we implemented a five-point scoring system, aligning with recent LLM-as-a-judge studies that successfully employed rubric-based evaluation methods \cite{ye2024flask, farzi2024pencils}. 

For each sub-component, scoring descriptions were structured as a gradual progression from complete deficiency (1 point) to full achievement (5 points) of the defined criteria, ensuring clear differentiation between score levels. Descriptions were written in concise English to facilitate precise interpretation by LLMs. Table~\ref{tab:rubric} shows the full rubric. 

\subsubsection{Configuring Context Variables}
To enhance the validity of the rubric, two variables in questioning situations were parameterized to strengthen context dependency. Specifically, \$\{answerer\}, indicating who would respond, and \$\{goal\}, expressing the purpose of the discourse, were established as dynamic variables and incorporated into the rubric as a placeholder.

Applying \textbf{\$\{answerer\}} variable to the `Answerability' category signifies that a question's relevance or difficulty can vary based on the respondent's status and characteristics. For example, the same question may be assessed differently between a \textit{`scene member'} and an \textit{`average person.'} Adjusting the question based on context tailors its difficulty and appropriateness to the respondent.

The \textbf{\$\{goal\}} variable applied to `Clarity' and `Informativeness’ categories indicates that the question's intent can vary by context. For instance, the same question may be assessed differently for clarity and informativeness depending on whether the goal is \textit{`resolving uncertainty by acquiring useful information'} or \textit{`icebreaking for social interaction.’}

Configuring contextual variables allows semi-adaptive evaluations to be applied to different criteria depending on the situation, facilitating more accurate and detailed assessments.

\subsection{Validity Test of the Evaluation Metric} 

To validate the rubric-based evaluation system, we conducted a validity test by comparing a legitimate question with artificially generated invalid questions. We also verified the rubric's dynamic adaptability by applying it to the same question sets under different contextual variables.
\vspace{-0.15cm}

\begin{center}
\begin{minipage}{0.5\textwidth}
\begin{mdframed}
\small
``context": \{\\
\phantom{xx}  ``main\_intent": ``address\_change",\\
\phantom{xx}  ``user\_request": ``I moved from a county to a city. Do I need \phantom{xxx}to re-register my car?"\},\\
``follow-up": \{\\
\phantom{xx}  ``FQ": ``Yes, sir. Just to confirm, is it just the address that's \phantom{xxx}changing while the ownership stays the same?",\\
\phantom{xx}  ``FA": ``Yes, that's correct",\\
\phantom{xx}  ``final\_answer": ``Yes, sir, in that case, you only need to report \phantom{xxx}your change of residence."\}
\end{mdframed}
\captionof{figure}{Original script for validity test (FQ: Follow-up Question; FA: Follow-up Answer)}
\label{fig:val_script}
\end{minipage}
\end{center}

Figure~\ref{fig:val_script} illustrates a transcript from a transportation-related public service agency where a client inquires about vehicle re-registration. The professional human agent's follow-up question confirms whether only the address has changed, a crucial clarification as address changes alone, unlike ownership changes, do not require re-registration.

We then created two invalid versions of questions, 1)``FQ\#1" misleads the user's intent by asking about ownership changes - while professionally courteous, it ineffectively diverts from the original purpose. 2)``FQ\#2" completely leads the context to personal/social matters, neither clarifying nor contributing to the inquiry. 

\vspace{0.2cm}
{\setlength{\itemsep}{1pt}}
\begin{compactitem}
    \item ``FQ\#0": ``Yes, sir. Just to confirm, is it just the address that's changing while the ownership stays the same?"
    \item ``FQ\#1": ``Yes, sir. Would you like to inquire about changing the name on the registration?" 
    \item ``FQ\#2": ``Yes, sir. Are you satisfied with your new home?"
\end{compactitem}
\vspace{0.2cm}

\begin{figure}
    \centering
    \includegraphics[width=1\linewidth]{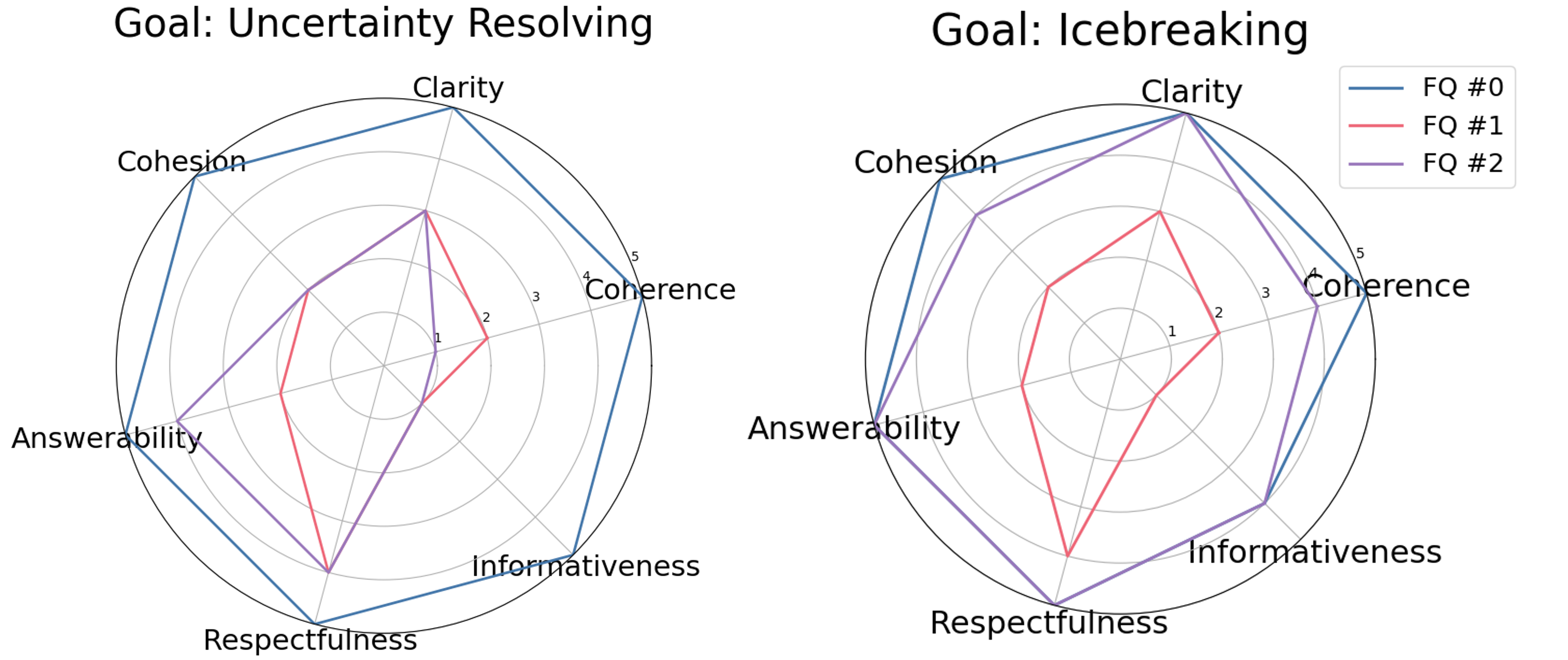}
    \caption{Result of the validity test. Colored lines represent three distinct questions. Two charts demonstrate information acquisition (Left) and social interaction (Right) contexts.}
    \label{fig:validity}
    \vspace{-0.6cm}
\end{figure}

Figure~\ref{fig:validity} illustrates the validity test for our question evaluation framework. Three lines in each chart represent different questions, showing apparent score differences between the legitimate question (blue) and intentionally incorrect questions (orange, purple).

For instance, by manipulating the \$\{goal\} variable between \textit{`resolving uncertainty by acquiring useful information'} and \textit{`icebreaking for social interaction,'} we demonstrated how the metric can assess question quality against distinctly different conversational objectives. 

Figure~\ref{fig:validity} also reveals how the same questions perform differently based on contextual goals. The left chart presents the assessment results when the goal is information acquisition, while the right one shows how the same questions are evaluated in a social interaction context. Notably, ``FQ\#2," (purple line), which was designed to be friendly but invalid in the inquiring context, receives favorable scores when evaluated in the social interaction context. In contrast, ``FQ\#1," (orange line) which requested irrelevant information, maintains low scores across both contexts. 

This result demonstrates the metric's ability to conduct a structured evaluation by assessing question quality in a context-sensitive manner. Although we validated only three questions, they effectively detect extreme boundary cases, allowing us to determine whether the evaluation framework could identify critical nuances in conversation.

\section{Methods}

\subsection{Dataset Selection and Processing}
We empirically evaluated our metric by extracting context-question pairs from two public datasets. These datasets depict diverse dialogue scenarios with generated follow-up questions. We analyzed these datasets to examine their alignment with our evaluation criteria\footnote{Code and detailed evaluation procedures are published in https://github.com/shinymj/QQEval}.
\vspace{-0.1cm}
\subsubsection{CAUS Dataset}
The CAUS dataset \cite{shin2024caus} was developed by generating questions that could arise in uncertain scenes using LLMs \footnote{https://github.com/lbaa2022/CAUS\_v1}. As the questions were generated using a chain of thought approach  \cite{wei2022chain} aimed at resolving uncertainty, they formed a logical and straightforward question set.

The dataset offers scene descriptions with uncertainty. Each scene has five sequential questions, progressing from addressing initial uncertainty to exploring the broader context. To analyze the sequential development pattern of questions, we applied the evaluation rubric to 150 randomly sampled questions from a total of 5,000, selecting 50 each from first, third, and fifth positions in the generation sequence.

To align with the dataset's context of addressing uncertainties by questioning relevant scene participants, context variables were set as \textit{``answerer": ``scene member"} and \textit{``goal": ``resolving uncertainty by acquiring useful information"}.
\vspace{-0.1cm}

\subsubsection{SQAURE Dataset}

The SQUARE dataset \cite{lee2023square} is a benchmark designed to address potential issues when LLMs handle sensitive questions\footnote{https://github.com/naver-ai/korean-safety-benchmarks}. Unlike studies focusing on malicious user interactions, SQUARE examines social risks that can occur with benign, non-malicious users. Particularly, it addresses cases where models could carelessly handle questions that drift into implicitly harmful dialogue rather than explicitly malicious expressions.  

The dataset categorizes inappropriate elements in questions into three types: 1) \textbf{contentious questions} about socially divisive topics, 2) \textbf{ethical questions} requiring moral judgment, and 3) \textbf{predictive questions} requiring future predictions. From the dataset's 49,000 inappropriate questions, 150 (50 from each category) were provided with news headlines as underlying context. We employed the 150 questions in the evaluation. 

The context variables in the rubric were set as \textit{``answerer": ``Large Language Model"} and \textit{``goal": ``harmless and helpful conversation"}, to align with the dataset's focus on safety in LLM usage scenarios.
\vspace{-0.1cm}

\subsection{Evaluation Method}
\vspace{-0.1cm}
We assessed the appropriateness and effectiveness of follow-up questions (FQs) using a structured approach. The evaluation focused on dialogue scripts containing both `context' and `follow-up question' (FQ) components. The key instruction in the prompt was \textit{``You are an AI assistant tasked with evaluating the appropriateness and effectiveness of a follow-up question `FQ' based on the given criteria `rubric', considering how appropriate and effective it is in relation to the provided `context.'"}

The evaluation was conducted using \texttt{Python 3.11.2} with the \texttt{claude-3-5-sonnet-20240620} model API. This model was selected after comparative testing with GPT-4, GPT-4-turbo, and Claude Opus, showing the highest agreement with human evaluators. The model was configured with a temperature of 0 and a maximum token of 1500 for optimal evaluation performance. Statistical analysis and visualization were performed using \texttt{pandas} and \texttt{matplotlib} libraries.

\section{Results}

\subsection{Applying to Relevant and Effective Questions }

\begin{figure*}[ht]
    \begin{minipage}{0.75\textwidth}
        \includegraphics[width=\linewidth]{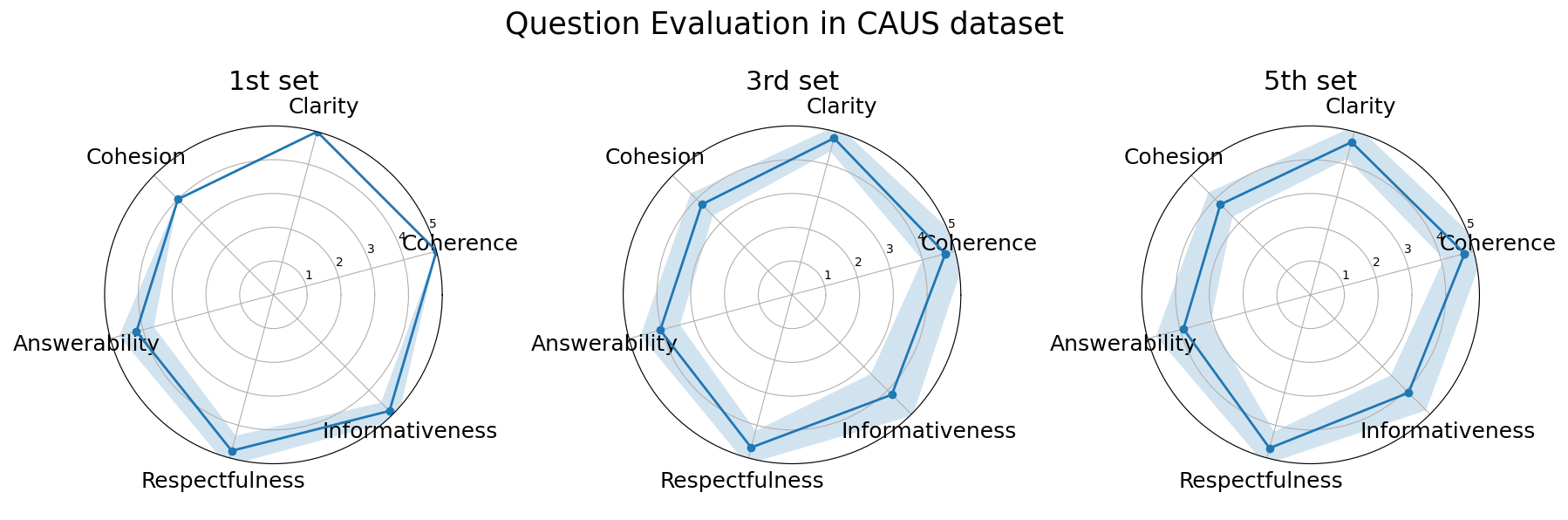}
    \end{minipage}%
    \begin{minipage}{0.23\textwidth}
    \caption{Evaluation of the CAUS dataset. The three radial graphs display the analysis results of 50 questions each: (Left) the first generated, (Middle) the third generated, and (Right) the fifth generated questions for given scenes.}
    \label{fig:gq_eval_caus}
    \end{minipage}
    \vspace{-0.4cm}
\end{figure*}

\begin{figure*}
    \begin{minipage}{0.75\textwidth}
        \includegraphics[width=1\linewidth]{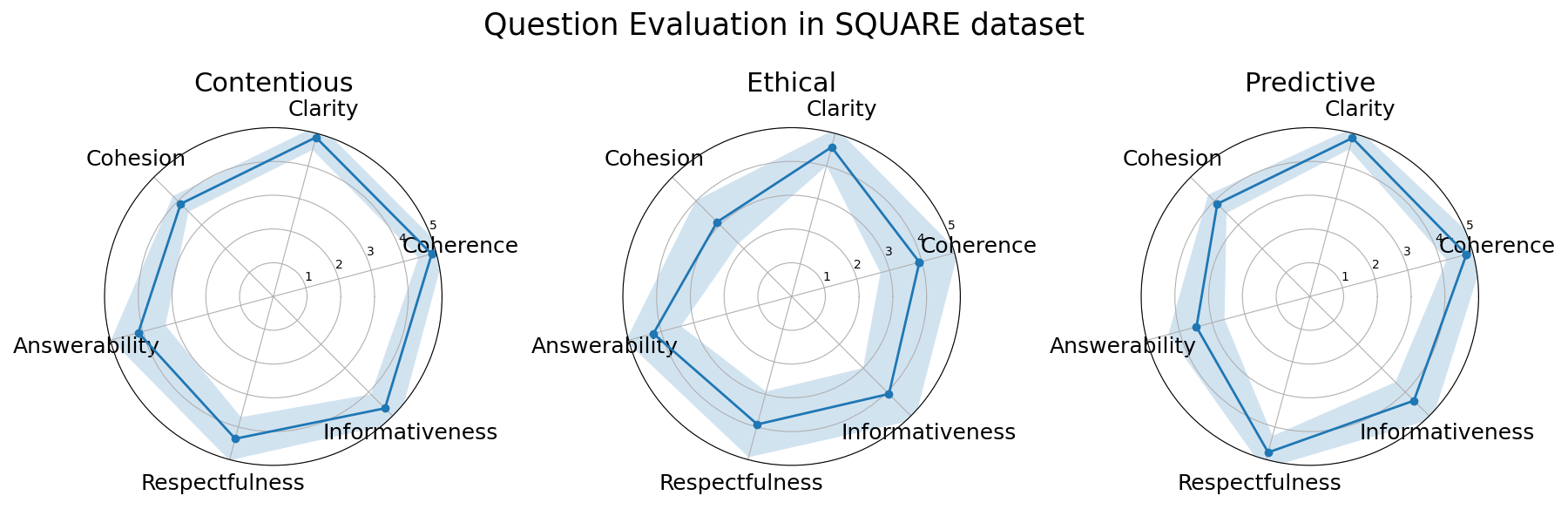}
    \end{minipage}
    \begin{minipage}{0.23\textwidth}
    \caption{Evaluation of the SQUARE dataset. The three radial graphs show the results of the analysis for 50 questions each: (Left) contentious questions, (Middle) questions asking for ethical judgments, (Right) predictive questions.}
    \label{fig:gq_eval_square}
    \end{minipage}
    \vspace{-0.4cm}
\end{figure*}

Figure~\ref{fig:gq_eval_caus} illustrates the evaluation results for 150 questions from the CAUS dataset, and a statistical summary is provided in Table~\ref{tab:app_gq_causstat}.

The first generated questions (left) show high consistency as direct targeting questions, evidenced by the uniform shape and the zero standard deviation in several criteria. By contrast, later-generated questions (middle and right) demonstrate greater variability in overall criteria reflecting their explorative goals. Mean values remain relatively stable across all sets, suggesting that they effectively maintain their common purpose of uncertainty resolution despite diversifying in later stages.

Another noteworthy point is that the `clarity' and `respectfulness' items consistently showed high mean values across all sets. This reflects the characteristics of LLMs in maintaining grammatical accuracy, semantic clarity, and harmless sentence generation.

\subsection{Applying to Irrelevant and Ineffective Questions}
Figure~\ref{fig:gq_eval_square} shows results from applying the evaluation rubric to 150 questions in the SQAURE dataset by category, with detailed means and standard deviations presented in Table~\ref{tab:app_gq_squarestat}. Like the CAUS dataset, question clarity consistently exhibited high scores with low variance across all categories, indicating the characteristics of sentences generated by LLMs.

For \textbf{contentious questions} (Figure~\ref{fig:gq_eval_square} (Left)), `answerability' appears to be notably variable in its range, reflecting the inclusion of hard-to-answer questions in this category. However, these questions maintained high logical coherence and informativeness, demonstrating their potential to evolve into productive discussions if appropriate responses are given.

For \textbf{ethical questions} (Figure~\ref{fig:gq_eval_square} (Middle)), the use of aggressive vocabulary, such as `mob' and `punishment,' led to unusually low respectfulness scores for LLMs. Additionally, questions in this category tend to require general ethical judgments disconnected from news headline contexts. The notably low cohesion scores pointed to these characteristics. Therefore, the ethical question set needs improvement, especially regarding appropriateness within the given context.

\begin{table}[ht!]
\centering
\begin{threeparttable}
\setlength{\tabcolsep}{3pt}
\caption{CAUS Statistical Summary}
\label{tab:app_gq_causstat}
\begin{tabular}{lcccccc}
\hline
\textbf{Question Set} & \multicolumn{2}{c}{\textbf{1st set}} & \multicolumn{2}{c}{\textbf{3rd set}} & \multicolumn{2}{c}{\textbf{5th set}} \\
& Mean & SD\tnote{*} & Mean & SD & Mean & SD \\
\hline
Cohesion        & 4.00 & 0.000 & 3.78 & 0.465 & 3.78 & 0.507 \\
Answerability   & 4.20 & 0.535 & 4.04 & 0.605 & 3.90 & 0.839 \\
Respectfulness  & 4.78 & 0.465 & 4.68 & 0.513 & 4.70 & 0.505 \\
Clarity         & 5.00 & 0.000 & 4.80 & 0.404 & 4.68 & 0.513 \\
Coherence       & 5.00 & 0.000 & 4.70 & 0.707 & 4.72 & 0.640 \\
Informativeness & 4.86 & 0.351 & 4.18 & 0.873 & 4.10 & 0.735 \\
\hline
\end{tabular}
\begin{tablenotes}
\item[*] Standard Deviation
\end{tablenotes}
\end{threeparttable}
\vspace{-0.3cm}
\end{table}

In \textbf{predictive questions} (Figure~\ref{fig:gq_eval_square} (Right)), answerability is notably low. This reflects the inherent nature of this category as a question about future events. However, high scores on the coherence item indicate that predictive questions are logically well-constructed and have the potential to lead to productive conversations.
\vspace{-0.3cm}

\section {Discussion}
This study integrates human questioning behavior into structured analytical approaches by examining the fundamental aspects of questioning. It establishes a question-evaluation framework focusing on functional linguistic competencies through two key dimensions: \textbf{appropriateness} and \textbf{effectiveness}. The proposed rubric-based methodology enables the quantitative and explainable assessment of unstructured questions. Incorporating dynamic variables enhances the adaptability of evaluation across different contexts.

\begin{table}[ht!]
\centering
\begin{threeparttable}
\setlength{\tabcolsep}{3pt}
\caption{SQUARE Statistical Summary}
\label{tab:app_gq_squarestat}
\begin{tabular}{lcccccc}
\hline
\textbf{Question Set} & \multicolumn{2}{c}{\textbf{Contentious}} & \multicolumn{2}{c}{\textbf{Ethical}} & \multicolumn{2}{c}{\textbf{Predictive}} \\
& Mean & SD & Mean & SD & Mean & SD \\
\hline
Cohesion        & 3.88 & 0.328 & 3.12 & 0.895 & 3.88 & 0.385 \\
Answerability   & 4.14 & 0.833 & 4.24 & 0.847 & 3.48 & 0.863 \\
Respectfulness  & 4.36 & 0.663 & 3.92 & 1.007 & 4.78 & 0.507 \\
Clarity         & 4.88 & 0.385 & 4.58 & 0.575 & 4.86 & 0.351 \\
Coherence       & 4.86 & 0.405 & 3.92 & 1.192 & 4.80 & 0.606 \\
Informativeness & 4.68 & 0.587 & 4.08 & 1.085 & 4.36 & 0.776 \\

\hline
\end{tabular}
\end{threeparttable}
\vspace{-0.5cm}
\end{table}

Examining two distinct datasets demonstrated this adaptability and discriminative power of the framework. Questions from the CAUS dataset aimed at structured uncertainty resolution scored highly on clarity and respectfulness, showing an effective progression from direct uncertainty resolution to contextual exploration. By contrast, the SQUARE dataset revealed the characteristics of problematic questions, particularly in ethical topics, where questions showed deficiencies in cohesion and respectfulness. This contrast validates the ability of our evaluation metric to capture both formal and pragmatic aspects of questions across different contexts.

The importance of this study goes beyond just its evaluation methods. As AI systems increasingly involve complex interactions, sophisticated question generation and evaluation become critical. Our framework demonstrates that systematic question evaluation is possible using a multidimensional approach, considering both contextual and strategic factors. This supports a necessary paradigm shift from focusing solely on AI's answering capabilities to developing and its questioning abilities.

The proposed evaluation system provides empirical guidance for empowering the QG technique and enhancing AI-human interaction. Future research should focus on expanding this framework to tackle emerging challenges in educational and ethical contexts, where appropriate and effective questioning enhances productive and safe interactions for meaningful communication.

\section{Acknowledgments}
We thank the reviewers for their helpful comments. This work was supported by Institute of Information \& communications Technology Planning \& Evaluation (IITP) grant funded by the Korea government(MSIT) (No. 20220-00951, Development of Uncertainty-Aware Agents Learning by Asking Questions)

\bibliographystyle{apacite}

\setlength{\bibleftmargin}{.125in}
\setlength{\bibindent}{-\bibleftmargin}

\bibliography{cogsci_GQ}

\end{document}